\documentclass{article}
\usepackage{amsmath}
\usepackage{algorithm}
\usepackage[noend]{algpseudocode}
\usepackage[pdftex]{graphics}

\makeatletter
\def\BState{\State\hskip-\ALG@thistlm}
\makeatother

\usepackage{kotex}
\usepackage{graphbox}
\usepackage{caption}
\usepackage{subcaption}
\usepackage{wrapfig}
\usepackage{xcolor}
\graphicspath{./figure/} 

\usepackage[numbers]{natbib}

\usepackage[final]{nips_2018}

\usepackage[utf8]{inputenc} 
\usepackage[T1]{fontenc}    
\usepackage{hyperref}       
\usepackage{url}            
\usepackage{booktabs}       
\usepackage{amsfonts}       
\usepackage{nicefrac}       
\usepackage{microtype}      

\makeatletter
\newcommand{\printfnsymbol}[1]{%
  \textsuperscript{\@fnsymbol{#1}}%
}
\makeatother

\title{Auto-Meta: \\
Automated Gradient Based Meta Learner Search}

\author{
  Jaehong Kim\textsuperscript{1}
  \And
  Sangyeul Lee\textsuperscript{1}
  \And
  Sungwan Kim\textsuperscript{1}
  \And
  Moonsu Cha\textsuperscript{1}
  \And
  Jung Kwon Lee\textsuperscript{1}
  \AND
  Youngduck Choi\textsuperscript{1,2}
  \And
  Yongseok Choi\textsuperscript{1}
  \And
  Dong-Yeon Cho\textsuperscript{1}
  \And
  Jiwon Kim\textsuperscript{1} 
  \AND
  \normalfont{SK T-Brain}\textsuperscript{1} \\
  Yale University\textsuperscript{2} \\
  \{ \texttt{xhark, sylee0335, sw0726.kim, ckanstnzja, jklee,} \\ 
  \texttt{yschoi, dycho24, jk}\} \texttt{@sktbrain.com} \\
  \texttt{youngduck.choi@yale.edu}
}

\begin{document}

\maketitle
\begin{abstract}
Fully automating machine learning pipelines is one of the key challenges of current artificial intelligence research, since practical machine learning often requires costly and time-consuming human-powered processes such as model design, algorithm development, and hyperparameter tuning. In this paper, we verify that automated architecture search synergizes with the effect of gradient-based meta learning. We adopt the progressive neural architecture search \cite{liu:pnas_google:DBLP:journals/corr/abs-1712-00559} to find optimal architectures for meta-learners. 
The gradient based meta-learner whose architecture was automatically found achieved state-of-the-art results on the 5-shot 5-way Mini-ImageNet classification problem with $74.65\%$ accuracy, which is $11.54\%$ improvement over the result obtained by the first gradient-based meta-learner called MAML \cite{finn:maml:DBLP:conf/icml/FinnAL17}.
To our best knowledge, this work is the first successful neural architecture search implementation in the context of meta learning.
 \end{abstract}
\section{Introduction}
Despite the lack of full knowledge of human learning mechanisms, many researchers tried to develop learning systems to mimic our ability to quickly adapt to new environments based on previous experiences. In \cite{DBLP:journals/corr/DuanSCBSA16:fast_rl,mishra:attentive_meta:DBLP:journals/corr/MishraRCA17,santoro:memoty_meta:DBLP:conf/icml/SantoroBBWL16,DBLP:journals/corr/WangKTSLMBKB16:learn_to_rl}, for example, recurrent neural networks (RNNs) were adopted to recognize the input and output mappings represented by training data and rapidly predict outputs for the test data using the internal states of learned RNN models. In contrast to these methods that rely on expert hand-crafted architectures, model-agnostic meta-learning (MAML) \cite{finn:maml:DBLP:conf/icml/FinnAL17,nichol:reptile:DBLP:journals/corr/abs-1803-02999} estimated a good initialization of model parameters for the fast adaptation to new tasks purely by a gradient-based search. Although we can apply model-agnostic meta-learning techniques to a variety of learning tasks without deeply contemplating the model architectures due to its model-agnostic properties, it is natural to expect that well-designed models for given tasks have better performances than conventional architectures.

In recent years, automation of machine learning has rapidly progressed. For instance, neural network architecture search
for image classification tasks has been successful. The selected architectures, when followed up with appropriate fine tuning, outperform models that are manually selected and trained by deep learning experts, a slow process that requires a large amount of trial and error guided by intuition.
Progressive neural architecture  search (PNAS) \cite{liu:pnas_google:DBLP:journals/corr/abs-1712-00559} 
is a particular form of the automated search that progressively expands the search candidate neural network architectures, supported by an RNN model for predicting candidates' performances without fully training the candidate architectures. Progressive neural architecture search achieves outstanding results without the significant computational expense required by reinforcement learning or evolutionary algorithm based searches.

In this paper, we propose a new approach to automatic design of neural network architectures for gradient-based meta learners. 
As the gradient-based meta-learning algorithm for the task, we considered Reptile \cite{nichol:reptile:DBLP:journals/corr/abs-1803-02999} which is an first-order approximation of MAML \cite{finn:maml:DBLP:conf/icml/FinnAL17}. To our best knowledge, this combination is the first instance of successful
scalable neural architecture search work within the meta learning literature. For 5-shot 5-way Mini-ImageNet classification problem, we obtained $74.65\%$ accuracy, which is $11.54\%$ improvement over the result from MAML \cite{finn:maml:DBLP:conf/icml/FinnAL17}.

\section{Related Works}

\paragraph{Meta-learning}


Meta learning as a theme is quite general, and extends well beyond the gradient based meta learners
for few shot classification tasks. In fact, much of the meta learning literature focuses on the general reinforcement
learning tasks \cite{DBLP:journals/corr/DuanSCBSA16:fast_rl, DBLP:journals/corr/WangKTSLMBKB16:learn_to_rl}.
One of the most common approaches to meta-learning is to build a recurrent neural network as a meta-learner. 
In particular, RNN based methods, augmented with
memory-augment network \cite{santoro:memoty_meta:DBLP:conf/icml/SantoroBBWL16} or simple attention mechanism \cite{mishra:attentive_meta:DBLP:journals/corr/MishraRCA17} have been
applied to few-shot image classification tasks. 

Metric learning is another popular approach to address meta-learning problems. The meta learner attempts to
learn a metric which can be used to compare two different examples effectively and performs tasks in the learned metric space \cite{Vinyals:nips2016:DBLP:conf/nips/VinyalsBLKW16}. Some studies train a Siamese network to achieve the same objective \cite{Koch:icmlw2015:DBLP:conf/icml/ShyamGD17}. The metric-based meta-learning has been known to perform well for few-shot image classification tasks \cite{Snell:nips2017:DBLP:conf/nips/SnellSZ17, Shyam:icml2017:DBLP:conf/icml/ShyamGD17}.

The other major category of meta learning is to learn an optimizer 
as the meta-learner which enables the learner to learn a new task more effectively \cite{Hochreiter:2001:DBLP:conf/icann/HochreiterYC01, Andrychowicz:nips2016:DBLP:conf/nips/AndrychowiczDCH16}. 
This approach has been applied to few-shot learning successfully \cite{ravi:opt_fewshot:DBLP:conf/iclr/RaviL17}. 
Rather than using the learned optimizer, a new meta-learning scheme applicable 
to all gradient-based learning algorithms, model-agnostic meta-learning (MAML), has recently been proposed \cite{finn:maml:DBLP:conf/icml/FinnAL17}. MAML attempts to find a set of parameters which initializes 
a learner for any specific task to be trained quickly only with small amount of data. 
Although this technique has shown the effectiveness for various few-shot learning problems including few-shot image classification and reinforcement learning, Hessian vector product calculation during training requires a 
large amount of computation. The first-order approximation algorithm has been proposed to avoid the Hessian computations in \cite{nichol:reptile:DBLP:journals/corr/abs-1803-02999}. Some advantages of gradient based meta learners have also
been discussed in \cite{finn:universality_maml:DBLP:/journals/corr/abs-1710-11622}.

\paragraph{Neural network architecture search}
Neural network architecture search (NAS) is a methodology to 
automatically find optimal neural network architectures for a given task. There are various 
types of NAS, such as reinforcement learning based NAS and evolutionary algorithm based NAS. 
Reinforcement learning based NAS includes REINFORCE \cite{zoph:rl_nas_google:DBLP:journals/corr/ZophL16}, 
Q-learning \cite{zhong:q_learning_nas:DBLP:journals/corr/abs-1708-05552, baker:rl_nas:DBLP:journals/corr/BakerGNR16}, 
and PPO-type algorithms \cite{zoph:nasnet_google:DBLP:journals/corr/ZophVSL17}. Evolutionary algorithm based NAS has extensively been explored in \cite{real:large_scale_nas_google:DBLP:conf/icml/RealMSSSTLK17,
miikkulainen:evolv_nas:DBLP:journals/corr/MiikkulainenLMR17, xie:genetic_cnn:DBLP:conf/iccv/XieY17, 
liu:hier_nas:DBLP:journals/corr/abs-1711-00436,
real:amoebanet_reg_nas:DBLP:journals/corr/abs-1802-01548}.
For example, AmoebaNet \cite{real:large_scale_nas_google:DBLP:conf/icml/RealMSSSTLK17} applies an evolutionary
algorithm to the same search space of NASNet and achieves state-of-the-art results on image classification tasks.
Other methods deploy various types of reasonable heuristics that attempt to reduce computational cost. 
This line of thinking is present in hypernetworks \cite{brock:smash:DBLP:journals/corr/abs-1708-05344}, co-evolving NEAT \cite{miikkulainen:evolv_nas:DBLP:journals/corr/MiikkulainenLMR17}, boosting \cite{cortes:adanet:DBLP:conf/icml/CortesGKMY17,  huang:boost_nas:DBLP:journals/corr/HuangALS17}, MCTS \cite{negrinho:deeparch:DBLP:journals/corr/NegrinhoG17}, early stopping of unpromising models \cite{baker:nas_perf_pred:DBLP:journals/corr/BakerGRN17}, and progressive neural architecture 
search (PNAS) \cite{liu:pnas_google:DBLP:journals/corr/abs-1712-00559}.
In particular, PNAS expands the search space incrementally from simple to complex so that it can search architectures in an efficient way without limiting the search to the space of fully-specified architectures.

\section{Auto-Meta}
Our goal is to automatically find the optimal network architecture for gradient-based meta-learners. Considering the loss function $\mathcal{L}$ for given tasks $j$ represented by training and test data sets $(D_{j}^{Tr},D_{j}^{Te})$, this can be formulated as follows:
\begin{eqnarray}
\min_{A,\theta} \sum_{j} \mathcal{L} \big (D_{j}^{Te}, U(D_{j}^{Tr},\theta;A) \big ),
\label{eq1}
\end{eqnarray}
where $A$ and $\theta$ are the neural network architecture and its parameters, respectively. $U$ denotes the computation of parameter updates using one or more gradient descent steps.
A natural and simply way to solve this problem is to minimize the loss $\mathcal{L}$ over parameters keeping the candidate architectures fixed. Then, based on some promising architectures, more complicated architectures are searched progressively. By repeating these two steps, we can obtain a good approximate solution to Equation (\ref{eq1}).

As the gradient-based meta-learning algorithm, we adopt Reptile \cite{nichol:reptile:DBLP:journals/corr/abs-1803-02999} which showed comparable performance with MAML \cite{finn:maml:DBLP:conf/icml/FinnAL17} on few-shot classification problems using only first-order gradient information. This allows us to avoid the time-consuming calculation of second-order derivatives.
As the network architecture search method, we consider the PNAS algorithm \cite{liu:pnas_google:DBLP:journals/corr/abs-1712-00559} where three layers (i.e., block, cell, and network) of abstraction for representing a neural network topology were defined. At most $B$ blocks which represent a combination operators applied to two inputs are included in a cell. This cell is then stacked a certain number of times to create a full CNN. During the architecture search, the cells ``progressively'' get more complicated by adding a block to themselves. Without expensive training procedure, the performance of each cell is evaluated with a surrogate predictor, such as LSTM to rank all expanded candidate cells. Then, CNNs with the top $K$ cells are trained and evaluated. We continue in this way until each cell has the maximum number of blocks.


\section{Experiments and Results}
To implement our gradient based meta-learner search, 
a distributed system that deploys and trains many neural networks
in a stable manner is required. We apply a Kubernetes system \cite{Kubernetes} that supports utilities to deploy over clusters of GPUs, in order to achieve the parallelization of training and testing top models in each iteration of block progression. 
We used the system with $112$ P$40$ GPUs for experiments described in this section. 
Our architecture search procedure on Mini-ImageNet dataset takes $24$ hours on average after parallelization.


\subsection{Few Shot Image Classification}\label{experiment_details}
To evaluate the automated architecture search algorithm for the gradient-based meta-learning, we applied our method to few-shot image classification problems. We used two benchmark datasets Omniglot \cite{Lake1332} and Mini-ImageNet \cite{Vinyals:nips2016:DBLP:conf/nips/VinyalsBLKW16}\cite{ravi:opt_fewshot:DBLP:conf/iclr/RaviL17}. In the Omniglot dataset, there are 1623 handwritten characters from 50 different alphabets. The Mini-ImageNet dataset consists of $60,000$ images ($84\times 84$  pixels) with $100$ classes.
In particular, we conducted the 5-way, 1-shot and 5-shot classification tasks through the proposed gradient based meta learner search.
Meta-test accuracy of meta learner was used as score for the LSTM predictor. Based on predicted scores from surrogate the LSTM, we trained 100 most promising cell candidates.
After search completed, we chose the cell structure with best score for final training. We use $3\times3$ convolution, $5\times5$ factorized convolution, identity, $3\times3$ average pooling, and $3\times3$ max pooling as operations for blocks. For training parameters of the network and surrogate LSTM predictor, we used Adam optimizer with learning rate $0.01$. Other hyperparameters used in our experiments are given in Table \ref{Tab:hyperparameters}.

 \begin{table}[h!]
 \centering
 \begin{tabular}{l|l|l}
 \hline
  Algorithm& \#Params &Accuracy\\
  \hline
  Ours ($F=10$) & 25.9k & \bf{97.44} $\pm$ \bf{0.07}\%\\
  Ours$^{Transduction}$ ($F=10$) & 25.9k & \bf{98.94} $\pm$ \bf{0.05}\%\\
  MAML$^{Transduction}$ \cite{finn:maml:DBLP:conf/icml/FinnAL17}& 112.0k & 98.7 $\pm$ 0.4\%\\
  1st-order MAML$^{Transduction}$ \cite{nichol:reptile:DBLP:journals/corr/abs-1803-02999}& 112.0k &98.3 $\pm$ 0.5\%\\
  Reptile \cite{nichol:reptile:DBLP:journals/corr/abs-1803-02999}& 113.2k & 95.39 $\pm$ 0.09\%\\
  Reptile$^{Transduction}$ \cite{nichol:reptile:DBLP:journals/corr/abs-1803-02999}& 113.2k & 97.68 $\pm$ 0.04\%\\
  \hline
  Matching-Nets \cite{Vinyals:nips2016:DBLP:conf/nips/VinyalsBLKW16}&-& 98.1\%\\
 Prototypical-Nets \cite{Snell:nips2017:DBLP:conf/nips/SnellSZ17}&111.9k&  98.8\%\\
 \hline
 \end{tabular}
 \\ \smallskip
 \caption{Results on Omniglot dataset for the 1-shot 5-way classification\\
 (\textit{F}: The number of filters in the first convolution layer, 
 \textit{Transduction}: Prediction at test time using batch normalization with batch mean and variance \cite{nichol:reptile:DBLP:journals/corr/abs-1803-02999})}
 \label{Tab:Omniglot}
 \end{table}
As shown in Table \ref{Tab:Omniglot}, our architecture search algorithm successfully found an excellent cell in 1-shot 5-way image classification for the Omniglot dataset so that the full convolution neural network (CNN) constructed with the cell outperformed the human-designed CNN trained with the Reptile \cite{nichol:reptile:DBLP:journals/corr/abs-1803-02999}. It is also worth pointing out that our CNN model has much smaller number of parameters than other ones.

\begin{table}[h]
\centering
\begin{tabular}{r|l|l|l|l|l}
\hline
\multicolumn{2}{c|}{ } & \multicolumn{2}{c|}{1-shot 5-way} & \multicolumn{2}{c}{5-shot 5-way} \\
 &Algorithm&\#Params&Accuracy&\#Params&Accuracy\\
 \hline
  \textit{small}&Ours ($F=10,10$)& 28k & \bf{49.58} $\pm$ \bf{0.20}\% &28k& \bf{65.09} $\pm$ \bf{0.24}\%\\
  setting&Ours$^{Transduction} (F=10,10)$& 28k & \bf{54.02} $\pm$ \bf{0.13}\%&28k&\bf{69.77} $\pm$ \bf{0.31\%}\\

  & MAML$^{Transduction}$ \cite{finn:maml:DBLP:conf/icml/FinnAL17}& 32.8k &  $48.70 \pm 1.84 \%$&32.8k& 63.11 $\pm$ 0.92\%\\
 & Reptile \cite{nichol:reptile:DBLP:journals/corr/abs-1803-02999}& 34.7k & $47.07 \pm 0.26\%$&34.7k& 62.74 $\pm$ 0.37\%\\
 & Reptile$^{Transduction}$ \cite{nichol:reptile:DBLP:journals/corr/abs-1803-02999}& 34.7k & 49.96 $\pm$ 0.32\%&34.7k& 65.99 $\pm$ 0.58\% \\

 \hline\hline
 \textit{large}& Ours ($F=12,12$) & 98.7k & \bf{51.16} $\pm$ \bf{0.17}\% &94.0k& \bf{69.18} $\pm$ \bf{0.14}\%\\
 setting & Ours$^{Transduction} (F=12,12)$& 98.7k & \bf{57.58} $\pm$ \bf{0.20}\% &94.0k& \bf{74.65} $\pm$ \bf{0.19}\%\\
 & Matching-Nets \cite{Vinyals:nips2016:DBLP:conf/nips/VinyalsBLKW16} & - & 43.60\% &-& 55.30\%\\
 & Ravi and Laroche \cite{ravi:opt_fewshot:DBLP:conf/iclr/RaviL17} & - & $43.40 \pm 0.77\%$ &-& 60.20 $\pm$ 0.71\%\\
 & Prototypical-Nets \cite{Snell:nips2017:DBLP:conf/nips/SnellSZ17} & 113.1k & $49.42 \pm 0.78\%$ &113.1k& 68.20 $\pm$ 0.66\%\\
 \hline
\end{tabular}
\\ \smallskip
\caption{Number of parameters and accuracy on Mini-ImageNet for 5-way, 1-shot and 5-shot classification tasks}
\label{Tab:MiniImage}
\end{table}

Results for Mini-ImageNet dataset are shown in Table \ref{Tab:MiniImage}. For fair comparison, we tried to make full CNN models with the searched cell have similar capacity to models adopted by previous work in \textit{small} setting. For the 1-shot and 5-way task, our method showed the comparable performance with others. However, the results clearly showed the superiority of the automated search for the 5-shot and 5-way task. Without restricting the model size in terms of the number of parameters, we constructed the best performing CNN with the best cell by appropriately choosing the value of \textit{F} in \textit{large} setting. For both 1-shot and 5-shot tasks, our model had the best accuracy. More specifically, our gradient based meta learner which have the automatically searched cell achieved $74.65\%$ accuracy on the 5-shot 5-way task. This implies that our approach not only drastically improved performance of other meta-learners but also can compete with state-of-the-art techniques such as \cite{Rusu2018} for few shot classification tasks. The best cell architectures found in these tasks were shown in Figure \ref{fig:best_cells}.

To investigate how our progressive network architecture search algorithm can find the best cells for few-shot image classification tasks, we observed the distribution of depths of the promising cells as the search progresses. As shown in Figure \ref{fig:depth_stats}, the distributions are moving toward deeper architectures in both settings. Surely, further empirical studies would be required to have a more conclusive remark on this phenomenon.
\section{Conclusion and Future Work}
Our gradient based meta learner search was motivated by the current trend in machine learning communities. There have been many attempts to automate the machine learning pipelines in different manner. We tried a simple, but natural combination of two automation techniques: neural architecture search and gradient-based meta-learning. To our best knowledge, this implementation is the first successful AutoML execution in the context of meta learning.

Our results indeed show that automated architecture search is beneficial for improvement of the performance of meta-learning algorithms which tried to find good initialization for different tasks. Our gradient based meta learners with automatically search architectures have much better results than other meta-learners with human-crafted models on some few shot image classification tasks and compatible results with state-of-the-art techniques which employed more sophisticated auxiliary components such as encoder and decoder networks for the tasks.

Future work includes exploring the effect of operators in our search method, and further empirical studies on the shape of cells in terms of depth and width for good performance. Also, various kinds of neural architecture search methods should be explored in the context of meta learning. For example, it would be quite interesting to see if the architecture and parameters can be jointly optimized by formulating the model search in a differentiable manner. The applicability of our method can be tested to other domains which require the fast-adaptation property.

\bibliographystyle{abbrv}
\bibliography{References}

\newpage
\section*{Appendix}

\vspace{-0.1in}
\subsection*{Architecture Search Hyperparameters}

\vspace{-0.1in}
\begin{table}[h!]
\renewcommand\thetable{A1}
\centering
\begin{tabular}{l | r}
\hline
\multicolumn{2}{c}{Progressive neural architecture search} \\
\hline
Max num blocks (B) &  5 \\
Num filters in first layer (F) & 4, 10, 12, 32 \\
Beam size (K) & 100 \\
Num times to unroll cell (N) & 0, 1 \\
Feature Scale Rate & 1, 2 \\
Surrogate predictor \cite{liu:pnas_google:DBLP:journals/corr/abs-1712-00559} & Cell size $= 100$, Number of layers $= 1$\\
\hline
\multicolumn{2}{c}{Gradient-based meta-learning} \\
\hline
Inner iterations & 8 \\
Inner-batch size & 10 \\
Inner Adam learning rate & 0.001, 0.01 \\
Meta-batch size & 5\\
Outer step size & 0.3, 1.0\\
Outer iterations & 1000  \\
\hline
\end{tabular}
\caption{Hyperparameters}
\label{Tab:hyperparameters}
\end{table}
\vspace{-0.1in}

\subsection*{Further Empirical Analysis}
\begin{figure}[h!]
\centering
\includegraphics[width=0.95\textwidth]{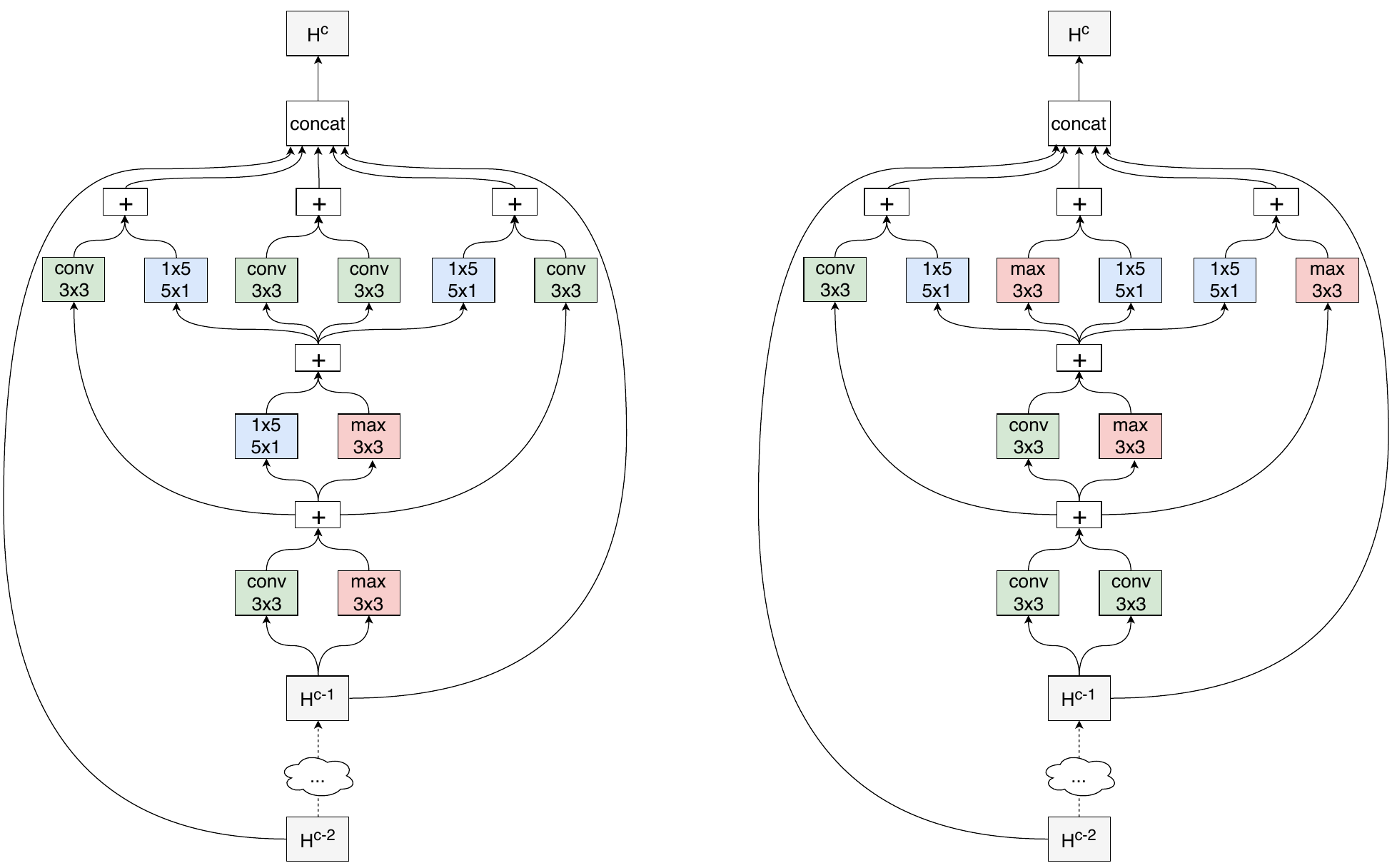}
\caption{The best cell architectures for 1-shot 5-way Mini-Imagenet tasks in the small (left) and large (right) settings (some pre-/post-processing operations are omitted for the sake of simplicity)}
\label{fig:best_cells}
\end{figure}

Figure \ref{fig:best_cells} shows the cell architectures that have the highest classification accuracies for 1-shot 5-way Mini-Imagenet classification tasks in the small and large settings respectively. The depths of both are equivalent to three vertically stacked blocks while their constituent operations and connections are different from each other.

\begin{figure}[h!]
	\begin{subfigure}[t]{0.49\textwidth}
	\includegraphics[width=\textwidth]{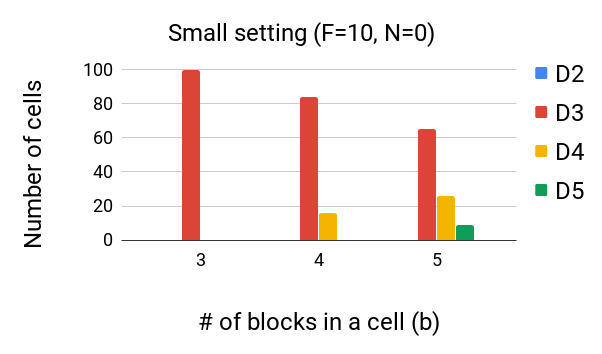}
	\end{subfigure}
	\begin{subfigure}[t]{0.49\textwidth}
	\includegraphics[width=\textwidth]{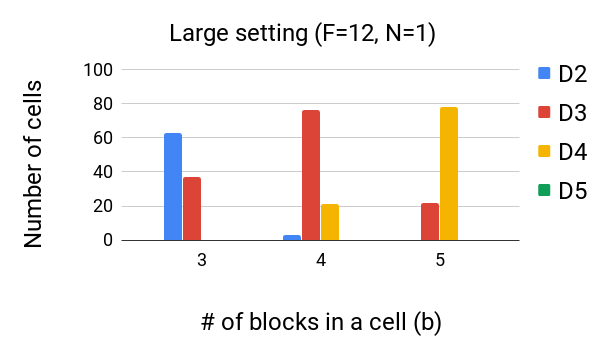}
	\end{subfigure}
        
\caption{Evolution of the cell depth distribution during the progressive network architecture search\\
         ($Dx$ refers to cells whose depth is $x$ in terms of blocks.)}
\label{fig:depth_stats}
\end{figure}

Figure \ref{fig:depth_stats} shows the distribution of depths of the best cells at each stage in terms of the number of blocks as the search progresses from $b=3$ to $b=5$. We can see the distributions are moving toward deeper architectures in both settings while the degree of the change seems much larger in the large setting.

\end{document}